\documentclass[letterpaper, 10 pt, conference, twoside]{ieeeconf}
\IEEEoverridecommandlockouts 

\let\labelindent\relax
\usepackage{graphicx}
\usepackage{eqlist}
\usepackage{amsfonts}
\usepackage{tabularx}
\usepackage{booktabs}
\usepackage{amssymb}
\usepackage{amsmath}
\usepackage{enumitem}
\usepackage{threeparttable}
\usepackage[lined, ruled, linesnumbered, commentsnumbered, noend]{algorithm2e}
\usepackage{lipsum}
\def\mathbi#1{\textbf{\em #1}}
\usepackage[usenames,dvipsnames]{xcolor}
\usepackage{comment}
\usepackage{flushend}

\usepackage{multirow}
\usepackage{multicol}
\usepackage[bottom]{footmisc}

\usepackage{geometry}
\newcommand{\squeezeup}{\vspace{-1.5mm}}

\title{\LARGE \bf Robotic Imitation of Human Assembly Skills Using Hybrid Trajectory and Force Learning}
\author{Yan Wang$^{1*}$, Cristian C. Beltran-Hernandez$^{1}$, Weiwei Wan$^{1}$ and Kensuke Harada$^{1,2}$
\thanks{*Correspond to:
{\tt\small yan@hlab.sys.es.osaka-u.ac.jp}}
\thanks{$^{1}$ Department of Systems Innovation, Graduate School of Engineering Science,
        Osaka University, Japan.}%
\thanks{$^{2}$ Automation Research Team, Artificial Intelligence Research Center, National Institute of Advanced Industrial Science and Technology, Japan.}%
}

\begin{document}


\maketitle

\begin{abstract}
Robotic assembly tasks involve complex and low-clearance insertion trajectories with varying contact forces at different stages.
While the nominal motion trajectory can be easily obtained from human demonstrations through kinesthetic teaching, teleoperation, simulation, among other methods, the force profile is harder to obtain especially when a real robot is unavailable. It is difficult to obtain a realistic force profile in simulation even with physics engines. Such simulated force profiles tend to be unsuitable for the actual robotic assembly due to the reality gap and uncertainty in the assembly process.
To address this problem, we present a combined learning-based framework to imitate human assembly skills through hybrid trajectory learning and force learning. 
The main contribution of this work is the development of a framework that combines hierarchical imitation learning, to learn the nominal motion trajectory, with a reinforcement learning-based force control scheme to learn an optimal force control policy, that can satisfy the nominal trajectory while adapting to the force requirement of the assembly task. To further improve the imitation learning part, we develop a hierarchical architecture, following the idea of goal-conditioned imitation learning, to generate the trajectory learning policy on the \textit{skill} level offline.
Through experimental validations, we corroborate that the proposed learning-based framework is robust to uncertainty in the assembly task, can generate high-quality trajectories, and can find suitable force control policies, which adapt to the task's force requirements more efficiently.

\end{abstract}

\section{Introduction}
\label{section:intro}

Industrial robots are usually expected to automatically complete the product assembly process. Consider an assembly task that has the following difficulties: (1) there are nonlinear and low-clearance insertion trajectories; (2) varying force control policies are needed at different phases. To reduce the risk of damage, lower contact forces are required in the search and early contact phase, while higher ones are desired in the insertion phase to guarantee stability and cope with friction. The assembly task requires skillful maneuvering and control. These difficulties make the task challenging for robots.

A promising solution to this problem is the use of imitation learning (IL) or learning from demonstration to reproduce movements observed from human demonstrations. Human demonstrations include trajectory profiles and force profiles. A trajectory profile can be obtained from human demonstration easily and numerously: with a real robot, we can collect it by kinesthetic teaching or teleoperation; when a real robot is unavailable, we can demonstrate the trajectory by observing the rendered world in simulation. However, in simulation, it is harder to obtain the realistic force profile due to the reality gap between simulation and real assembly.
Moreover, assembly tasks involve considerable task uncertainty, mainly including (1) trajectory uncertainty, i.e. motion drift of the robot end effector (EEF), and (2) force uncertainty, i.e. the varying accompanying forces of changing trajectory phases. Therefore, the improper force profile significantly impairs the implementation of such assembly tasks and can even damage parts. For these reasons, approaches that can execute proper force control along with the trajectory with high adaptivity to uncertainty must be explored to solve this problem.

One line of such researches is the belief space planning \cite{bry2011rapidly}, which means planning under state uncertainty. For robotic assembly tasks, \cite{wirnshofer2018robust} introduces an asymptotically optimal belief space planner that can find an optimized trajectory which has achieved minimal contact forces under uncertainty. However, such approach requires CAD models of manipulated objects to generate sets of pose hypotheses of them in the environment.
Another approach that has also been widely studied recently by researchers is reinforcement learning (RL) to solve complex manipulation problems, by learning behaviors through interacting with the environment.
Among the previously proposed RL techniques, hierarchical RL (HRL) \cite{barto2003recent} has been proven to be very promising in solving the above-mentioned uncertainty problems, although the method is problematic in exploration, skill segmentation, and reward definition \cite{gupta2019relay}. IL can be used to reduce the enormous exploration work of RL but is often influenced by compounding errors caused by the covariate shift \cite{ross2010efficient}\cite{ross2011reduction} and suboptimal expert demonstrations \cite{cheng2018fast}. 
\begin{figure}[t]
      \centering
      \setlength{\belowcaptionskip}{-5pt}
      \includegraphics[width=0.95\linewidth]{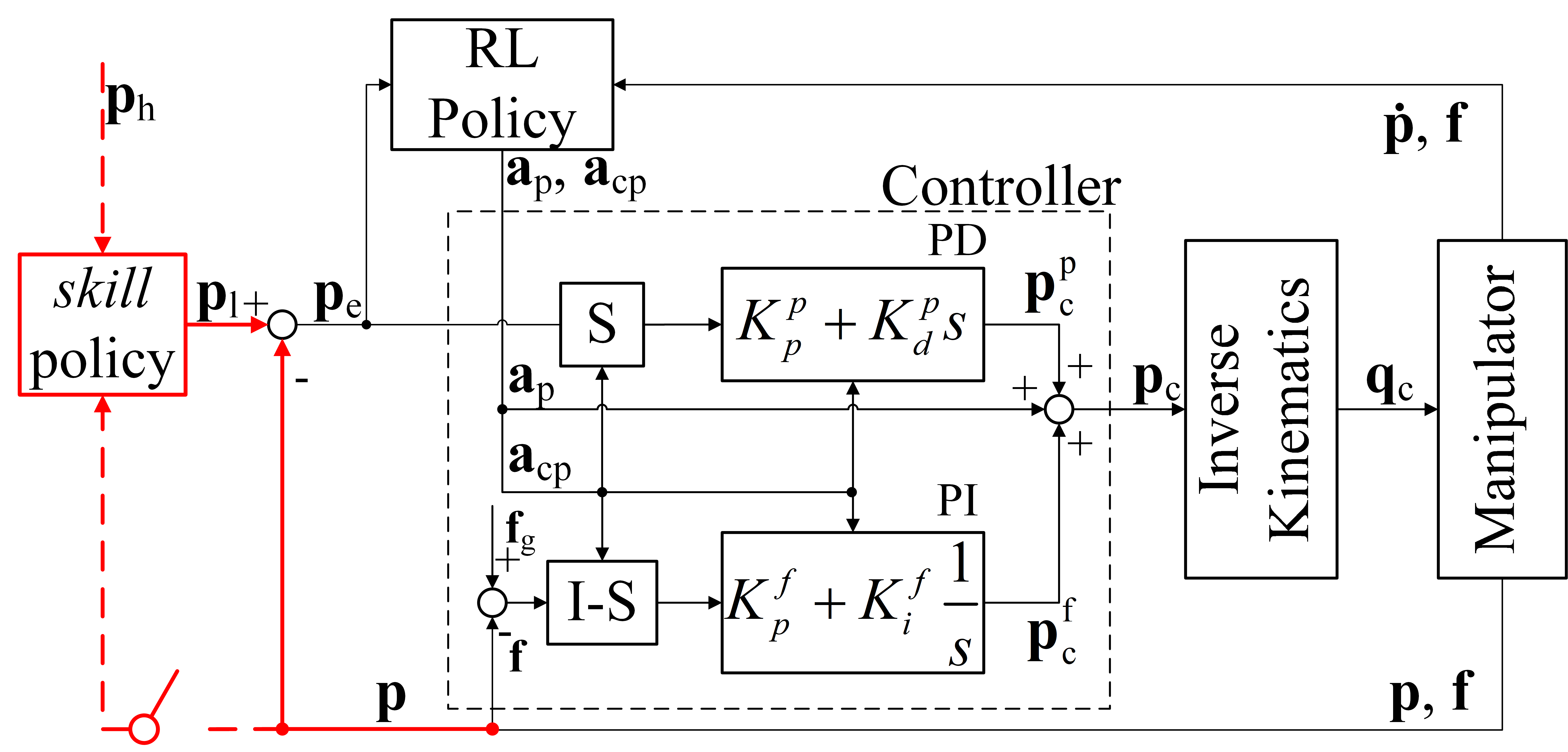}
      \caption{Combined learning (CL) framework. In the control loop, the \textit{skill} policy takes the goal pose $\mathbf{p}_h$ and the current pose $\mathbf{p}$ as inputs, and generates a new sub-goal $\mathbf{p}_l$ every certain time steps (represented by the switch symbol). Parallel position/force controller: a PD controller generates part of the motion trajectory, $\mathbf{p}_c^p$, based on $\mathbf{p}_e$ and a PI controller adjusts the motion trajectory by commands $\mathbf{p}_c^f$ according to the force $\mathbf{f}$. $K_p^p$ = PD proportional gain, $K_d^p$ = PD derivative gain, $K_p^f$ = PI proportional gain, $K_i^f$ = PI integral gain.}
      \label{fig:rlscheme}
   \end{figure}

In this work, we propose a learning-based framework to enable robots to independently and efficiently learn the proper force control policy of assembly tasks that compensates for the discrepancy between the force profile and the actual force requirement. The main contribution is the development of the said learning-based framework that combines hierarchical imitation learning, to learn the nominal motion trajectories, with a reinforcement learning-based force control scheme to learn an optimal force control policy, that can satisfy the nominal trajectory while adapting to the force requirements of the assembly task.
To further improve the imitation learning part, we develop a hierarchical architecture, following the idea of goal-conditioned imitation learning, to generate the trajectory learning policy on the \textit{skill} level offline. Through experimental validations, we corroborate that our framework is robust to uncertainty in the assembly task, can generate high-quality trajectories, and can find suitable force control policies adapting to the task's force requirements more efficiently.
\section{Related Work}
\label{section:relatedwork}
\subsection{Force Control in Assembly Tasks}
Force control is important for contact-rich manipulation tasks to deal with the problem of interaction between a manipulator and the environment.
Robotic assembly tasks involve nonlinear and low-clearance insertions and different force requirements in different phases. These make assembly automation difficult to achieve.
In conventional force control methods, prior knowledge of the environment is necessary to define the controller's parameters at each phase properly \cite{hogan1984impedance}. 
Therefore, it is tricky to adapt to unexpected changes in the environment and large forces can arise from small tolerances in assembly processes.
To overcome this problem, some typical previous studies involve extracting features from demonstrations and formulating numerical models to reuse these skills. 
For example, Dynamical Movement Primitives (DMPs) \cite{ijspeert2013dynamical} are presented for learning and modeling with attractor dynamical systems for goal-directed behavior. And researchers \cite{kruger2014technologies}\cite{abu2015adaptation}\cite{savarimuthu2017teaching} utilize the DMPs to encode the demonstrated trajectory and match the force profile of the recorded trajectory with the force profile of the current trajectory. \cite{Roveda2017iterative} proposes an iterative learning controller with reinforcement for tracking the target force with high accuracy in robotized tasks.

Since these approaches treat forces from human demonstrations as optimal solutions, they directly match the force profiles.
In contrast, we consider the situation where the force profiles are unsuitable or unavailable for the actual robotic assembly and propose a method that directly learns the time-variant parameters of the force controller from the experience of implementing assembly tasks.
\subsection{Combined Learning for Assembly Skills}
With the development of hardware and machine learning, a recent trend has emerged of using RL for complex assembly tasks, which circumvents complex numerical modeling.
Inoue \textit{et al.} \cite{inoue2017deep} propose a deep RL (DRL) method for precise PiH tasks.
Thomas \textit{et al.} \cite{thomas2018learning} solve contact-rich manipulation problems by incorporating CAD-based motion planning with RL.
Johannink \textit{et al.} \cite{johannink2019residual} combine feedback control methods with DRL to solve complex manipulation tasks involving friction and contacts with unstable objects.
However, the sample efficiency of RL can be rather low in exploring the task-space of assembly tasks.
IL is also a powerful method for robotic manipulation that perceives and reproduces human movements without the need for programming. Kormushev \textit{et al.} \cite{kormushev2011imitation} propose an IL approach that regulates the relative priority of position- and force-tracking for the controller in tasks with less uncertainty.
IL has also been applied recently to robotic assembly by researchers \cite{takamatsu2007recognizing}\cite{tang2016teach}\cite{suomalainen2017geometric}.
Previous works have also studied combined learning (CL) approaches, i.e. combining RL with IL. These approaches utilize demonstrations of successful behaviors to improve the sample efficiency of RL \cite{atkeson1997robot}.
Initializing RL policies from demonstration has been used for learning some classical tasks such as cart-pole \cite{atkeson1997robot}, hitting a baseball \cite{peters2008reinforcement}, and swing-up \cite{kober2009policy}.
Beyond initialization, some promising approaches incorporate demonstrations into RL, including DDPGfD \cite{vecerik2017leveraging}, DDPG-HER \cite{nair2018overcoming}, and DAPG \cite{rajeswaran2017learning}. 
However, demonstrations without proper force profiles are not suitable to initialize the RL policies for the force learning or to be incorporated into the RL process in these approaches.
Therefore, we present a novel CL framework that can learn a proper force control policy with high sample efficiency using only the trajectory profile.

\section{Preliminaries}
\label{section:preliminaries}
\subsection{Goal-Conditioned Imitation Learning}
\label{section:goalconditioned}
IL learns a policy by imitating a demonstration dataset $\mathcal{D} = \left\{\tau^1,\tau^2,\tau^3,...\right\}$ which consists of several trajectories on the same task. In a typical IL setting, a demonstration trajectory is in the form of state-action pairs, i.e. $\tau^i = (s^i_0,a^i_0,...,s^i_T,a^i_T)$, and a policy $\pi(a|s)$ is learned by behavior cloning (BC) \cite{bain1995framework}. 
In contrast, in a goal-conditioned IL setting \cite{kaelbling1993learning}\cite{schaul2015universal}\cite{ding2019goal}, state-action-goal triplets, $(s^i_{t}, a^i_{t}, s^i_{g})$, replace state-action pairs, $(s^i_t,a^i_t)$, and a goal-conditioned policy, $\pi(a|s,s_g)$, attempting to reach different goals $s_g$ is learned by imitating $\mathcal{D}$.
\subsection{Data Relabeling}
\label{section:datarelabel}
Data relabeling \cite{lynch2019learning} is a type of data augmentation that is particularly effective in a low data regime. 
It executes self-supervision on unlabeled demonstration data and treats each state $s^i_{t+k}$ visited within a demonstration trajectory, from an initial state $s^i_t$ to a specific goal state $s^i_g$, as a latent goal state. The effectiveness of data relabeling has been validated in \cite{ding2019goal} through a comparison with BC.
\section{Combined Learning Framework}
\label{section:method}
In this section, we propose our framework as shown in Fig.\ref{fig:rlscheme}, including the trajectory learning and the force learning.
\subsection{Hierarchical Goal-Conditioned Imitation Learning}
\label{section:hgcil}
We develop a hierarchical goal-conditioned IL (HGCIL) approach for the trajectory learning. 
The details are as follows:
\subsubsection{Hierarchical Architecture}
\label{section:bilevel}
\begin{figure}[t]
      \centering
      \setlength{\belowcaptionskip}{-12pt}
      \includegraphics[width=0.95\linewidth]{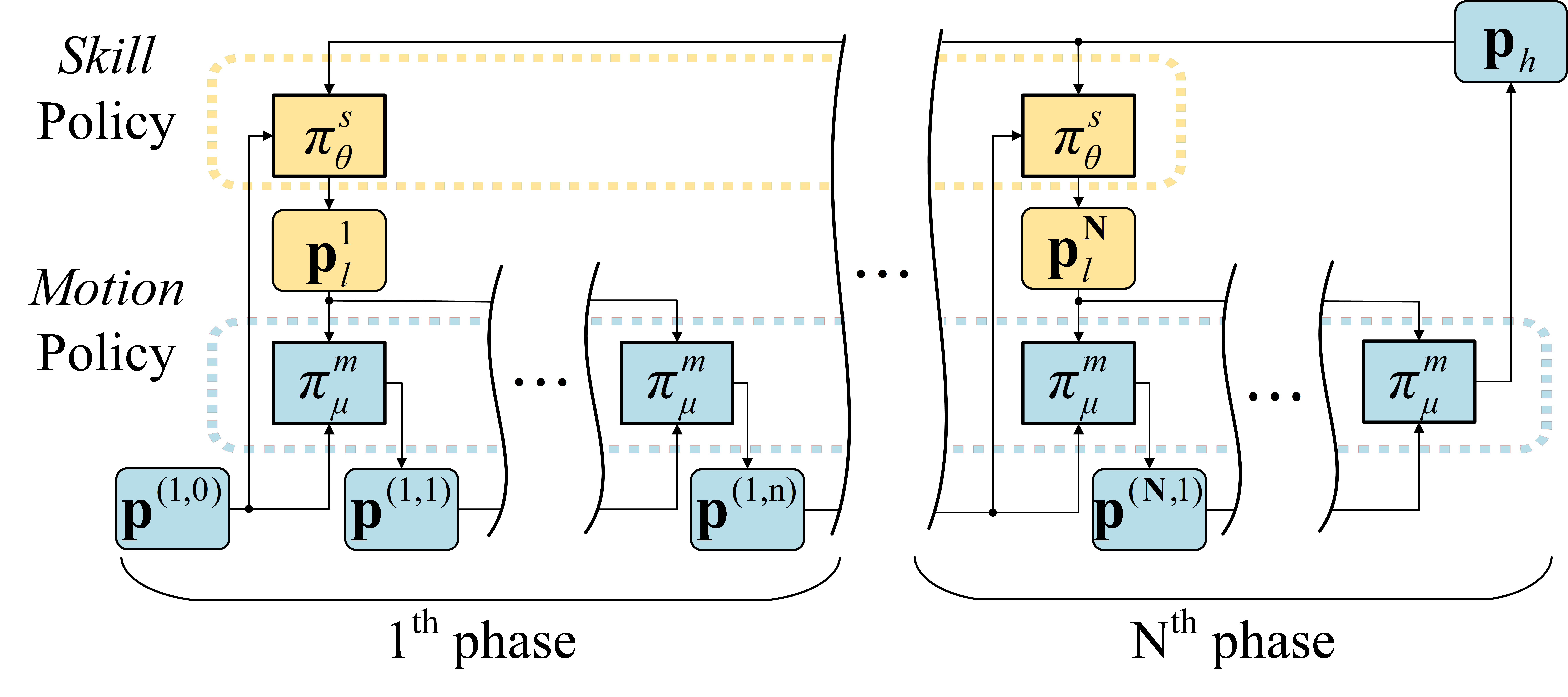}
      \caption{The hierarchical architecture of the trajectory learning. The \textit{skill} level policy sets a sub-goal, $\mathbf{p}_l$, every $\mathrm{n}$ time steps and commands the \textit{motion} level to execute actions to reach $\mathbf{p}_l$ at each phase until achieving the goal $\mathbf{p}_h$ at the $\mathrm{N^{th}}$ phase.}
      \label{fig:bilevel}
   \end{figure}

As the basic technique of HGCIL, we propose a hierarchical architecture consisting of a \textit{skill} level and a \textit{motion} level. We separate this part from Fig.\ref{fig:rlscheme} (red lines) and show it in temporal sequence in Fig.\ref{fig:bilevel}.

On the \textit{skill} level, the observation space is $(\mathbf{p}, \mathbf{p}_h)\in \mathbb{R}^m$ where $\mathbf{p}$ represents the current pose of the manipulated object and $\mathbf{p}_h$ stands for the goal pose on the \textit{skill} level.
The action space is $\mathbf{p}_l\in \mathbb{R}^n$ where $\mathbf{p}_l$ is the sub-goal pose on the \textit{motion} level.
And on the \textit{motion} level, the observation space is $(\mathbf{p}, \mathbf{p}_l)\in \mathbb{R}^m$ where $\mathbf{p}_l$ stands for the sub-goal pose obtained from the \textit{skill} level.
The action space is $\mathbf{a}_p \in \mathbb{R}^k$ where $\mathbf{a}_p$ is the robot EEF movement transitioning from $\mathbf{p}$ to $\mathbf{p}_l$.
During the trajectory learning process as shown in Fig.\ref{fig:bilevel}, the \textit{skill} policy sets a sub-goal $\mathbf{p}_l$ every $\mathrm{n}$ time steps and keeps it during each phase for the \textit{motion} level. Then, the \textit{motion} level is commanded to reach $\mathbf{p}_l$ every time step throughout the phase until a new $\mathbf{p}_l$ is given. 
\subsubsection{Goal-Conditioned Data Relabeling}
The goal-conditioned setting has shown good performance in multitasks and long-horizon tasks. In this paper, we use it to preprocess demonstration data. 
Data relabeling is implemented on the \textit{skill} level to generate the \textit{skill} level dataset $\mathcal{D}_s = \left\{S | S=(\mathbf{p}, \mathbf{p}_l, \mathbf{p}_h)\right\}$ as shown in Algorithm \ref{algorithm:relabel}. 
\begin{algorithm}[b]
\SetAlgoLined
\SetAlCapFnt{\scriptsize}
Initialization: input demonstration dataset $\mathcal{D} = \left\{\tau^1,\tau^2,\tau^3,...\right\}$, the window size $W_s$ for \textit{skill} level, and $W_m$ for \textit{motion} level\;
\For{each trajectory $\tau=(\mathbf{p}^0,\mathbf{p}^1,...,\mathbf{p}^T)$ in $\mathcal{D}$}{
 \For{$t=1,2,...,T$}{
  \For{$w=1,2,...,W_s$}{
    \If {$t+w$$\leq$$T$}{
        Add $(\mathbf{p}^t,\mathbf{p}^{t+min(w,W_m)},\mathbf{p}^{t+w})$ to $\mathcal{D}_s$\;}
  }
 }
}
\caption{Hierarchical goal-conditioned data relabeling}
\label{algorithm:relabel}
\end{algorithm}
\subsubsection{Policy}
On the \textit{skill} level, the \textit{skill} policy takes $(\mathbf{p}, \mathbf{p}_h)$ as inputs and outputs $\mathbf{p}_l$. We use a fully connected neural network with three hidden layers, each with 256 units, a dropout rate of 0.1, and ReLu as the activation function to train the \textit{skill} policy, $\pi^s_\theta(\mathbf{p}_l|\mathbf{p},\mathbf{p}_h)$. We use input feature normalization because the input data have quite different ranges and units. The output data is not normalized because it is a regression problem.
On the \textit{motion} level, the policy takes $(\mathbf{p}, \mathbf{p}_l)$ as inputs and outputs $\mathbf{a}_p$ every time step.
We incorporate this part with the RL-based force control scheme which will be presented in Section \ref{section:rlscheme}. 

We summarize how these techniques make HGCIL work here.
First, the hierarchical architecture endows the \textit{skill} policy with the capability to spontaneously find sub-goals for a distant goal along the trajectory. With the periodically updated $\mathbf{p}_l$, the motion drift uncertainty caused by the compounding errors can be constrained. Second, the goal-conditioned data relabeling enables the \textit{skill} policy to be trained on abundant goals in the trajectory distribution from the varying demonstration. Therefore, the \textit{skill} policy has a better generalization to move towards the goal from unseen states.
\subsection{Reinforcement Learning-Based Force Control Scheme}
\label{section:rlscheme}
For the force learning, we use an RL-based force control scheme to learn the force control policy. We continue to use the notation of the object pose in Section \ref{section:hgcil} to represent the EEF pose in this section because they can be transformed to each other with a known transformation matrix.

\subsubsection{Scheme}
The RL-based force control scheme is presented with black lines in Fig.\ref{fig:rlscheme}. The goal of this framework is to self-tune force control parameters without changing them manually according to the environment.
   
The goal pose of the manipulator EEF, which is $\mathbf{p}_l$ in Fig.\ref{fig:rlscheme}, is input into the scheme as a trajectory reference every time step. $\mathbf{p}$ is the actual pose of the EEF.
The pose error, $\mathbf{p}_e = \mathbf{p}_l-\mathbf{p}$,
and the contact force, $\mathbf{f}=[\,\mathbi{f},\boldsymbol{\tau}]$, where $\mathbi{f}\in\mathbb{R}^3$ is the force vector and $\boldsymbol{\tau}\in\mathbb{R}^3$ is the torque vector, serve as feedback to both the RL agent and the parallel position/force controller.
The velocity of the EEF, $\dot{\mathbf{p}}$, is also an input of the RL agent.
The RL agent gives policy actions consisting of (1) position/orientation commands,
$\mathbf{a}_p=[\mathbi{v},\mathbi{w}]$,
where $\mathbi{v}\in\mathbb{R}^3$ is the position vector and $\mathbi{w}\in\mathbb{R}^4$ is the orientation vector to control the robot's movements, and (2) controller parameters, $\mathbf{a}_{cp}$. The controller processes all inputs to produce the position command actually to be sent to the manipulator.

\subsubsection{Controller}
We utilize an adaptive parallel position/force controller to balance the accurate trajectory and the contact forces. The controller takes the trajectory error $\mathbf{p}_e$, the actual contact force $\mathbf{f}$, position/orientation commands $\mathbf{a}_p$, and controller parameters $\mathbf{a}_{cp}$ as inputs and produces the actual position command $\mathbf{p}_c$ to operate the manipulator; $\mathbf{f}_g$ is the reference force in the assembly process.
The parallel position/force controller control law consists of a proportional and derivative (PD) controller for position, a proportional and integral (PI) controller for force, and the selection matrix $S=diag(s_1,s_2,...,s_6),s_j\in[0,1]$
where the values correspond to the degree of control that each controller has over a given direction. 
The controller parameters are $\mathbf{a}_{cp}=[K_p^p,K_p^f,S]$.
Each parameter is defined in a range of potential values and can be self-tuned in the RL.

\subsubsection{Reinforcement Learning Algorithm}
Soft-Actor-Critic (SAC) \cite{haarnoja2018soft} is used as the RL algorithm of the RL agent in the scheme. SAC is a state-of-the-art model-free and off-policy actor-critic deep RL algorithm based on the maximum entropy RL framework. It maximizes the reward to an expected value while optimizing maximum entropy. As an off-policy algorithm, it can use a replay buffer to reuse information from recent operations for sample-efficient training. We use the SAC implementation from TF2RL \cite{ota2020tf2rl} repository.
The reward function is:
\begin{equation}\label{eq:reward}
    \begin{aligned}
        r(\mathbf{s}) = w_1L(\left\|\frac{\mathbf{p}_e}{\mathbf{p}_{max}}\right\|_{1,2})+w_2L(\left\|\frac{\mathbf{f}_e}{\mathbf{f}_{max}}\right\|_2)+\gamma
    \end{aligned}
\end{equation}
where $\mathbf{f}_e=\mathbf{f}_g-\mathbf{f}$, and $\mathbf{p}_{max}$ and $\mathbf{f}_{max}$ are defined maximum values.
$L(y) = y\rightarrow x, x\in[1,0]$ is a linear mapping to the range 1 to 0. Therefore, the closer the EEF is to the target and the lower the contact force is, the higher the reward. $\left\|z\right\|_{1,2}$ is $\mathit{l}_{12}$ norm based on \cite{levine2016end}, which is given by
$\frac{1}{2}\left\|z\right\|^2+\sqrt{\alpha+z^2}$. This norm is used to encourage the EEF to precisely reach the target position, but to also receive a larger penalty when far away.
$\gamma$ can be a positive reward for finishing the task successfully, a negative one for excessive force, or 0. $w_1$ and $w_2$ are hyperparameters to weight the components.

To deal with safety problem in the learning process, a fail-safe mechanism, which has been introduced in our previous work \cite{beltran2020learning}, checks the IK solution, the joint velocity, and the contact force in succession for each action. The first two validations are proactive to avoid unstable behaviors and the third one is reactive to avoid force overshoots.
\squeezeup
\section{Experiments}
\label{section:experiment}
\begin{figure}[t]
      \centering
      \setlength{\belowcaptionskip}{-5pt}
      \includegraphics[width=.8\linewidth]{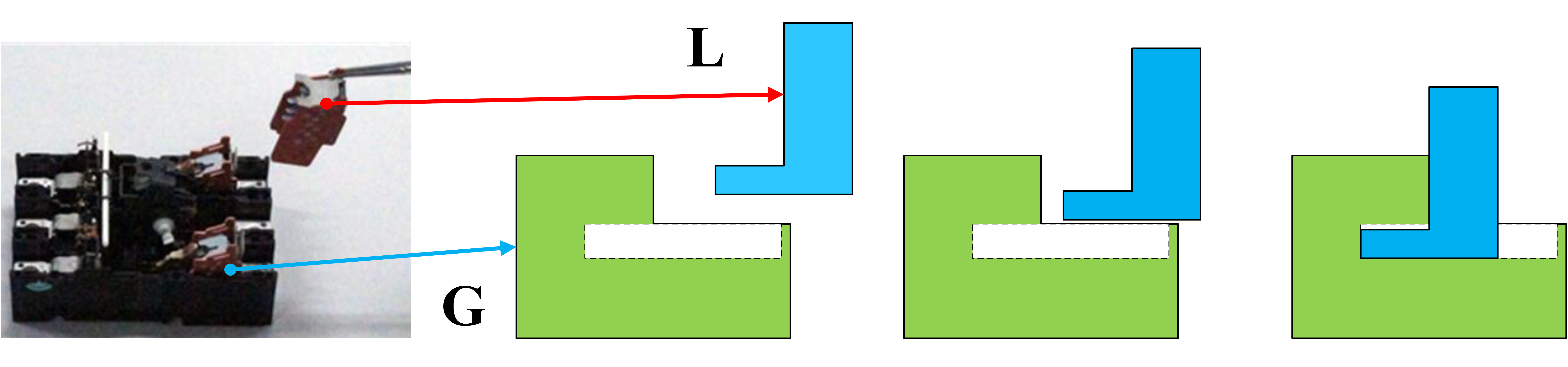}
      \ \ \ (a)
      \includegraphics[width=.85\linewidth]{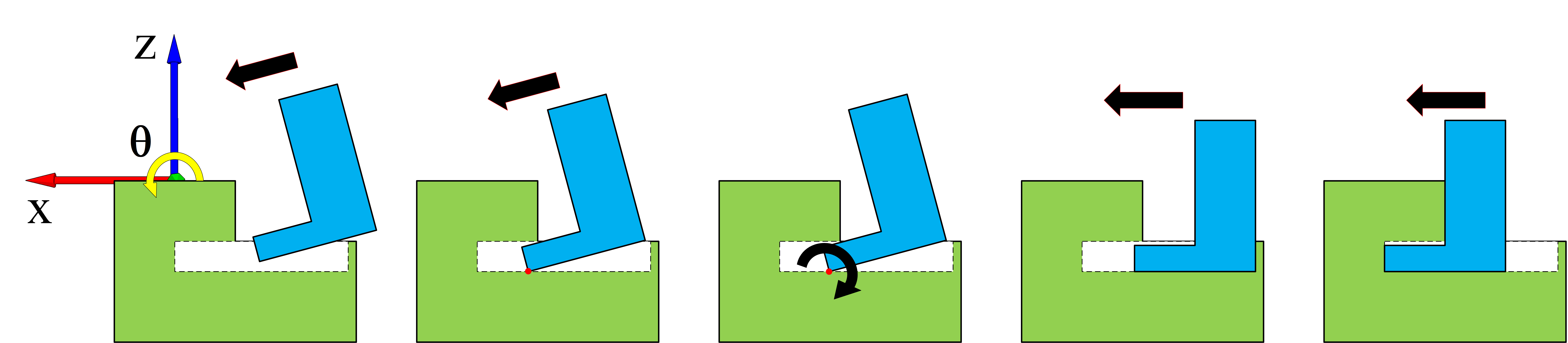}
      (b)
      \caption{\textit{L insertion} task. Clearances no more than 1 mm. (a) shows the circuit breaker condenser assembly can be simplified to an \textit{L insertion} task; (b) shows the assembly workflow of \textit{L insertion}.}
      \label{fig:breakerLG}
   \end{figure}

We divide this section into three parts: introducing the implementation details, explaining the choice of parameters, and evaluating the efficacy of the proposed framework on both the simulated environment and the real hardware.  
To improve the clarity, we explain with the circuit breaker condenser assembly which can be simplified into an L-shaped object insertion (\textit{L insertion}) task, as shown in Fig.\ref{fig:breakerLG}.
\squeezeup
\subsection{Implementation Details}
\subsubsection{Demonstration Device}
We collected data using two methods: (1) a bilateral control system constructed on a Yaskawa Motoman-SDA10F dual-arm robot; (2) AR markers attached to the manually manipulated objects.
For the former, we only used the trajectory data, albeit the force data is also available. The system recorded poses of the EEF every 4 ms. We refer readers to the media attachment for more information about the bilateral control system. For the latter, we tracked AR markers and recorded poses every 10 ms. 
\subsubsection{Setup for Experimental Validation}
Experimental validation was executed both on a simulated environment modeled in the Gazebo 9 and on a real Universal Robot 3e-series (UR3e) robotic arm. 
The real UR3e robotic arm had a control frequency up to 500 Hz, and a force/torque sensor was attached to the EEF. The learning process was implemented on a computer with a GeForce RTX 2060 SUPER GPU and an
Intel Core i7-9700 CPU.
\subsection{Choice of Parameters}
\subsubsection{Trajectory Learning}
The \textit{skill} level dataset, $\mathcal{D}_s$, was organized according to Algorithm \ref{algorithm:relabel}.
The object pose at time $t$, $\mathbf{p}^t$, was represented by the task-space pose, ($\mathrm{x}_t$, $\mathrm{y}_t$, $\mathrm{z}_t$, $\mathrm{\phi}_t$, $\mathrm{\theta}_t$, $\mathrm{\psi}_t$). 
We utilize task-space position/orientation commands and parallel position/force controller parameters as the motion representation for training because task-space actions have been proven to be more efficient for learning with RL algorithms \cite{bellegarda2019training}\cite{martin2019variable}.
Because our demonstration data lay in ranges far from the discontinuity, the trained neural network would not suffer from the non-continuous orientation representation.
Therefore, we used the Euler angle representation of the orientation for convenience. 
The neural network of the \textit{skill} policy was trained with the ADAM optimizer using a batch size of 256 and a learning rate decaying from 0.001. The window sizes $W_s$ and $W_m$ were set to 85\% and 30\% of the length of each trajectory, respectively. We used an original demonstration dataset including 10 trajectories (8 for the training set and 2 for the validation set).
\subsubsection{Force Learning}
There were 30 parameters in total in the parallel position/force controller, 12 from the PD gains, 12 from the PI gains, and 6 from the selection matrix $S$.
We reduced the number of controllable parameters to guarantee the behavior stability and to reduce the complexity of the system. In the PD controller, only the proportional gain $K_p^p$ was controllable and the derivative gain $K_d^p$ was computed with $K_p^p$. We set $K_d^p = 2\sqrt{K_p^p}$ to have a critical damping condition.
In the PI controller, only the proportional gain $K_p^f$ was controllable and the integral gain $K_i^f$ was computed based on $K_p^f$. We set $K_i^f$ as 1\% of $K_p^f$ empirically. 
To overcome the curse of dimensionality in RL, it was still necessary to further reduce the dimensions of parameters. In the experiment, we evaluated several policy models with different action spaces, i.e. different numbers of controllable parameters \cite{beltran2020learning}. As shown in Table \ref{table:actionspace}, we named models with different action spaces P-9, P-14, P-19, and P-24, where P meant \textit{parallel} and the numbers indicated the numbers of controllable parameters.
\renewcommand{\arraystretch}{1.5}
\setlength{\tabcolsep}{0.5pt}
\begin{table}[t]
\centering
    \caption{Policy Models with Different Action Spaces}
    \resizebox{0.95\linewidth}{!}{
    \begin{tabular}{ m{2.5cm}<{\centering} m{1.5cm}<{\centering} m{1.5cm}<{\centering} m{1.5cm}<{\centering} m{1.5cm}<{\centering}}
    \hline\hline
    \multirow{3}{*}{\textbf{Model}} & \multicolumn{4}{c}{\textbf{Number of Controllable Parameters}}   \\ \cline{2-5} 
                                                          & $\mathbf{a}_p$      & \multicolumn{3}{c}{$\mathbf{a}_{cp}$} \\ \cline{2-5} 
                                                           & Pose    & PD     & PI     & S    \\ 
    \hline
    P-9                           & 6       & 1      & 1      & 1    \\
    P-14                          & 6       & 1      & 1      & 6    \\
    P-19                          & 6       & 6      & 6      & 1    \\
    P-24                          & 6       & 6      & 6      & 6    \\ 
    \hline\hline
    \end{tabular}
    }
    \label{table:actionspace}
\end{table}
\squeezeup\squeezeup\squeezeup\squeezeup
\subsection{Experimental Results}
We implemented several experiments on both the simulation and the real hardware. The \textit{skill} policy set $\mathbf{p}_l$ every 50 time steps (10\% of the time steps per episode) empirically to balance the computational resource and the trajectory quality.
\subsubsection{Simulation Experiments }
\begin{figure}[t]
      \centering
      \setlength{\belowcaptionskip}{-5pt}
      \includegraphics[width=1.0\linewidth]{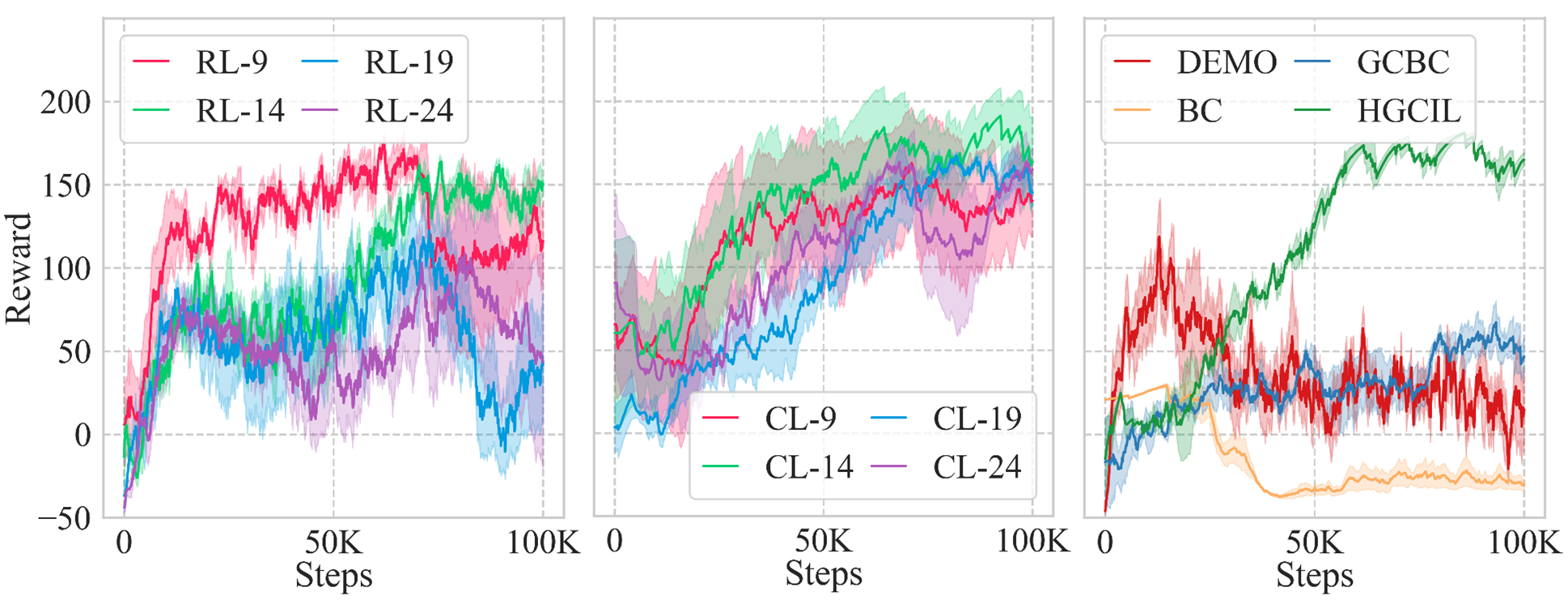}
      \caption{ Learning curves of training sessions. Left: RL; Middle: CL; Right: Different learning frameworks.}
      \label{fig:learningcurve}
   \end{figure}

In this part, we implemented all the simulation experiments on the \textit{L insertion} task.
First, we trained each policy model with both the CL framework and the vanilla RL to verify whether the trajectory learning helped to improve the sample efficiency. Each policy model was trained for 100K steps with a maximum of 500 steps per episode. The policy control frequency was set to 20 Hz, so each time step cost about 0.05 s in the simulation. 
Parameters in the reward function (Eq.\ref{eq:reward}) were: $w_1$ = 1.0, $w_2$ = 2.0, and $\gamma$ = 100 (reaching the goal), -50 (excessive force), or 0 (otherwise). The reference force, $\mathbf{f}_g$, was set to 0 to minimize the contact force.
The learning curves smoothed using the exponential moving averages (EMA) are shown in Fig.{\ref{fig:learningcurve}}.

\textit{Result 1:} The learning curves of the vanilla RL hardly converged, but a learning curve could also converge with proper tuning of the hyperparameters, such as P-14. However, the learning curves of the CL converged more frequently without the need for careful parameter tuning.
Among the policy models trained with CL, P-14 showed the fastest learning rate and the best performance. 
P-19 and P-24 had higher system complexity with more parameters to tune than the other two, so their learning rates were lower. P-9 had a fast learning rate but lacked enough controllable parameters to learn a good policy consistently.

Second, to evaluate the performance of HGCIL in imitating the nominal motion trajectory confronted with the trajectory uncertainty, we compared it with two other IL approaches: (1) Behavior cloning (BC) without hierarchical or goal-conditioned setting, and (2) GCBC with flat data relabeling (GCBC-relabeling) \cite{lynch2019learning}. The window size of GCBC-relabeling equals $W_s$ used by HGCIL. And the learning of \textit{motion} policy of each approach was still incorporated into the force control scheme. For comparison, we additionally included a framework where the same force control scheme followed a recorded trajectory from the demonstration.
Therefore, we obtained four comparison groups, including HGCIL+SAC (ours), BC+SAC, GCBC-relabeling+SAC, and recorded trajectory+SAC. We set \textbf{L} to a random initial pose in the proximity of \textbf{G} and executed each learning process. The results are shown in Fig.\ref{fig:learningcurve}.
   
\textit{Result 2:} On the one hand, HGCIL+SAC outperformed BC+SAC and GCBC-relabeling+SAC. It indicated that the ability of HGCIL to recover from slight trajectory deviations contributed to adapt to the trajectory uncertainty. On the other hand, BC+SAC failed to learn proper policy throughout the training process; GCBC-relabeling improved the performance of the approach which merely followed a recorded trajectory, but was less effective to learn proper force control policy than HGCIL.
It suggested that these two methods were not robust enough for the trajectory learning and delivered inferior reference trajectories for the force learning.

\begin{figure}[t]
      \centering
      \setlength{\belowcaptionskip}{-5pt}
      \includegraphics[width=0.85\linewidth]{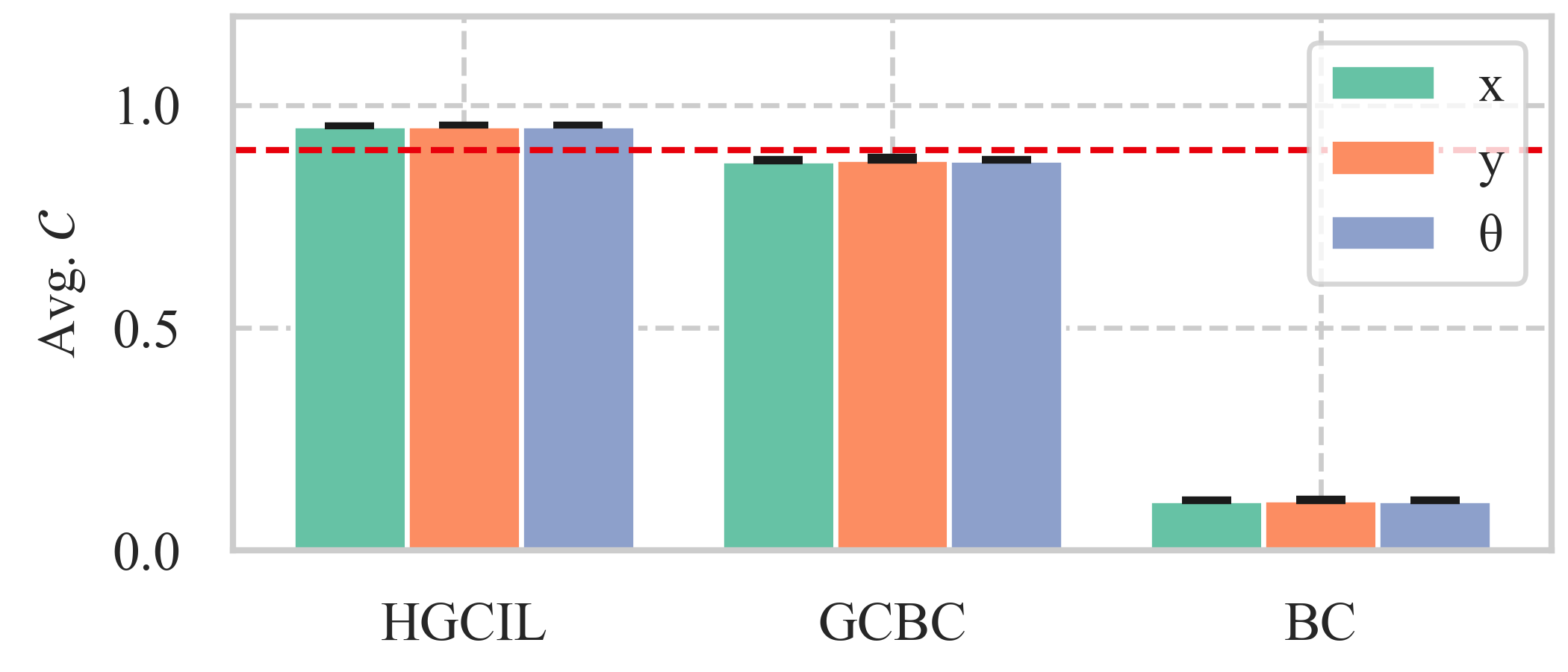}
      \caption{Score of different IL approaches. Starting poses of \textbf{L} are sampled from a uniform distribution centered at the initial pose of a demonstration trajectory with variance of $\pm10mm$, $\pm10mm$, and $\pm6^\circ$ in three main DoF, $\mathrm{x}$, $\mathrm{z}$, and $\mathrm{\theta}$, respectively, of the task-space within the close proximity of \textbf{G}.}
      \label{fig:score}
   \end{figure}

\begin{figure}[t]
      \centering
      \setlength{\belowcaptionskip}{-5pt}
      \includegraphics[width=0.9\linewidth]{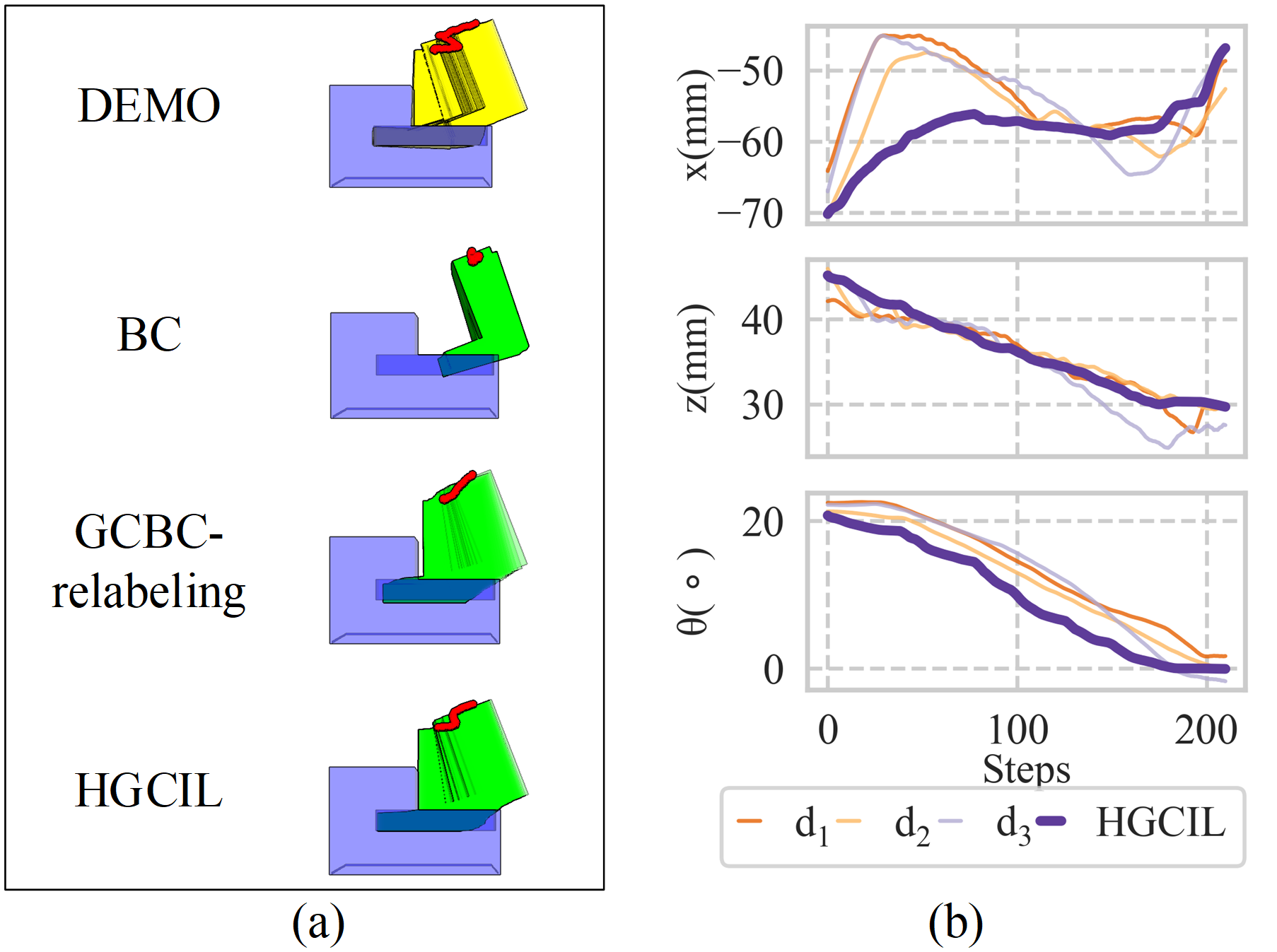}
      \caption{Trajectory learning results. (a) A demonstration trajectory (DEMO) and learned trajectories by HGCIL, BC, and GCBC-relabeling from near initial poses. Red trajectories represent trajectories of the local coordinate system origin of \textbf{L}. Yellow and green areas represent planar swept volumes of \textbf{L} along the trajectories. (b) A learned trajectory by HGCIL and 3 demonstration trajectories of the \textit{L insertion} are shown for the 3 main DoF - $\mathrm{x}$, $\mathrm{z}$, and $\mathrm{\theta}$.}
      \label{fig:ILpaths}
   \end{figure}

Third, to display the robustness of HGCIL more clearly, as well as validate the brief summary about its working mechanism in Section \ref{section:hgcil}, we excluded the contact force from this comparison and solely validated the trajectory learning part.
The performance was measured by a score $\mathcal{C}$ representing the extent of completion of the \textit{L insertion}, $\mathcal{C}=\mathrm{1-\lambda_{pos}e_{pos}-\lambda_{rot}e_{rot}}$, where $\mathrm{e_{pos}}$ and $\mathrm{e_{rot}}$ are the final Euclidean distances between the EEF and the target pose in position and orientation. $\mathrm{\lambda_{pos}}$ and $\mathrm{\lambda_{rot}}$ are their weights, which were set to 20.0 and 3.0, respectively. 
We used different starting poses of \textbf{L} uniformly distributed in three main degrees of freedom (DoF), $\mathrm{x}$, $\mathrm{z}$, and $\mathrm{\theta}$, of the task-space within the close proximity of \textbf{G} as comparisons. 20 trials were executed for each test group, and a trial finished when $\mathcal{C}$ surpassed a value of 0.9 or the implementation exceeded 500 time steps.
The result is shown in Fig.\ref{fig:score}.
More intuitively, Fig.\ref{fig:ILpaths}a gives the visible trajectory and planar swept volume of \textbf{L} obtained by each IL method with the initial starting pose. We emphasize that the object \textbf{G} in Fig.\ref{fig:ILpaths}a is only a position reference to show the extent of completion of the trajectory learning and the collision detection was inactive.

\textit{Result 3:} From Fig.\ref{fig:score} and Fig.\ref{fig:ILpaths}a, we found that both HGCIL and GCBC-relabeling succeeded in moving towards the goal from varying initial poses. It manifested the generalization ability of the goal-conditioned data relabeling to deal with situations not provided by demonstration. And we analyzed that the trajectory learning using BC stopped making progress early because BC could not produce proper actions under unseen states without the goal-conditioned data relabeling. 
However, the trajectory learned through GCBC-relabeling deviated more than that learned through HGCIL. 
It was because the \textit{skill} policy of HGCIL spontaneously found and updated sub-goals for a distant goal.
Therefore, the trajectory was always corrected to avoid serious deviation and the motion drift uncertainty was constrained. 
We also compared the trajectory learned by HGCIL with some demonstration trajectories (DEMO). As shown in Fig.\ref{fig:ILpaths}b, the trajectory learned by HGCIL was more consistent than the suboptimal demonstration trajectories and approximately an optimal trajectory reaching the goal without redundant movements (e.g. a demonstration trajectory may fluctuate a lot to adjust the object pose because of the improper insertion angle), which is also indicated in Fig.\ref{fig:ILpaths}a.
\subsubsection{Real Robot Experiments}
In this part, we executed three assembly tasks on real hardware. The policy control frequency was set at 20 Hz on the real robot.

\textit{Task 1} was a real \textit{L insertion} task as a testbed to select the policy model for \textit{Task 2} and \textit{Task 3}.
\begin{figure}[t]
      \centering
      \setlength{\belowcaptionskip}{-5pt}
      \includegraphics[width=1.0\linewidth]{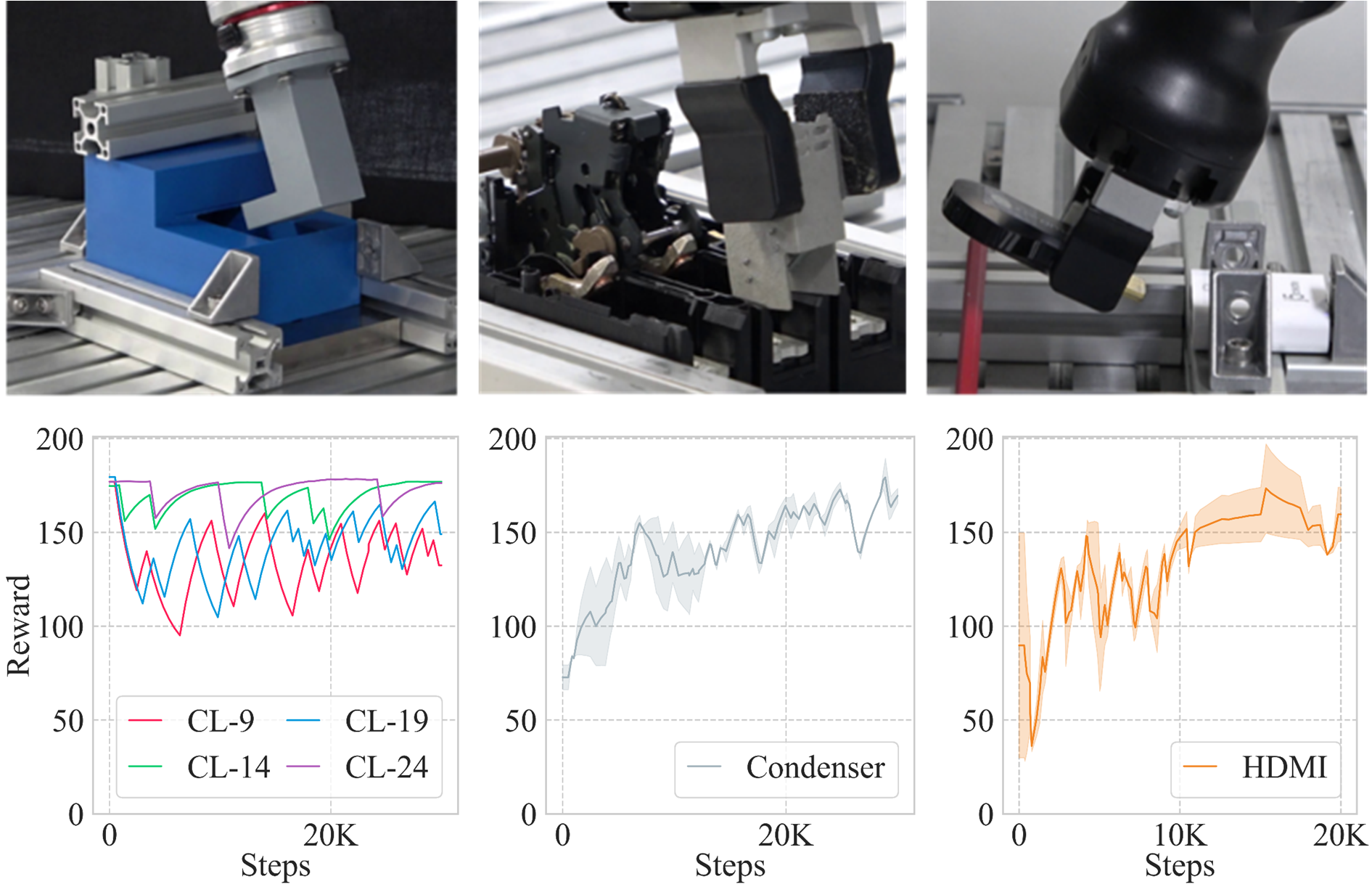}
      \caption{Left: \textit{L insertion} task, 30K-step training sessions of different policy models trained with CL; Middle: Condenser assembly, 30K-step training session; Right: HDMI insertion, 20K-step training session.}
      \label{fig:realassembly}
   \end{figure}
Then, we utilized \textit{Task 2} and \textit{Task 3} to validate that the CL framework could easily learn proper control policy confronted with variable trajectories and force requirements in different assembly tasks. \textit{Task 2} was a condenser assembly task - the origin of the \textit{L insertion} task,
as shown in Fig.\ref{fig:realassembly}. Compared to the simplified \textit{L insertion}, it involved higher trajectory uncertainty due to the complicated geometric features. 
\textit{Task 3} was an HDMI insertion (with an obstacle) task, as shown in Fig.\ref{fig:realassembly}. Besides the nonlinear and low-clearance trajectory, this task requires a certain force against the friction to finish the insertion, which brought the force uncertainty. Therefore, we set the reference force in the insertion direction to 10 N. For trajectory learning, we collected trajectory data from their demonstrations. And for the force learning, we used the learned force control policy in the real \textit{L insertion} as initialization to reduce the training time.

\textit{Results:} 
For \textit{Task 1}, we continually trained the policies learned in the simulation environment on a real UR3e robotic arm and evaluated the finally learned policies, as shown in Fig.\ref{fig:realassembly}. In the training process, the model P-14 maintained the best performance. Although it took about 40K steps (33.3 min) to learn a proper policy using P-14 in the simulation, the policy was transferred to the real hardware successfully without retuning from scratch.
Therefore, we utilized P-14 as the policy model for the next two tasks.
For \textit{Task 2} and \textit{Task 3}, the results are also shown in Fig.\ref{fig:realassembly}.

\begin{figure}[t]
      \centering
      \setlength{\belowcaptionskip}{-5pt}
      \includegraphics[width=0.85\linewidth]{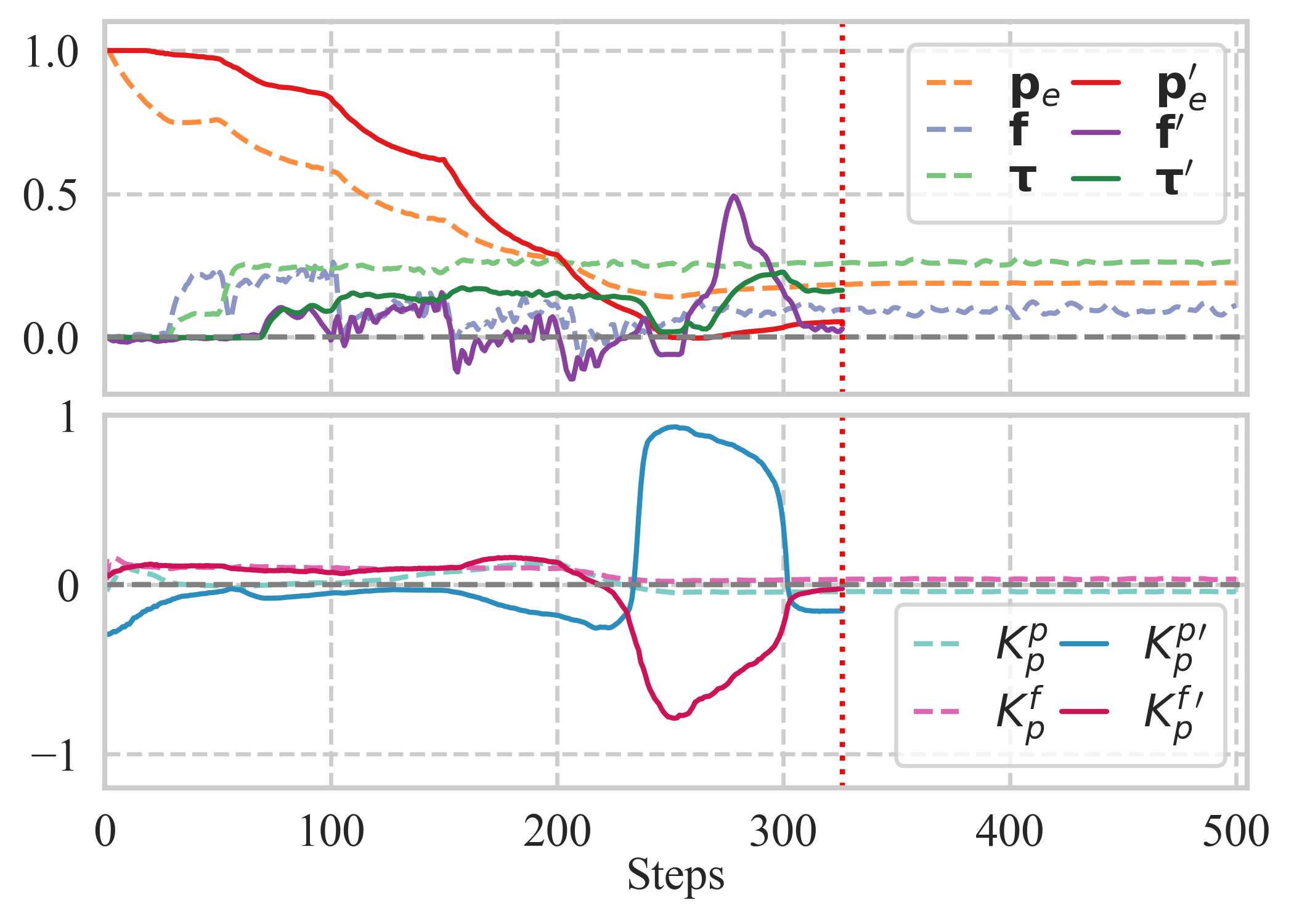}
      \caption{Policy evolution in the HDMI insertion task. $\mathbf{p}_e$, $\mathbi{f}$, and {$\boldsymbol{\tau}$} values have been mapped to a range of [1, -1]. The figures show the performances of the initial (dashed lines) and the learned (solid lines) policy across an evaluation session.}
      \label{fig:evolution}
   \end{figure}

To further explain how the controller adapted to the force uncertainty, i.e. the varying force requirements accompanying the trajectory profiles, the policy evolution of \textit{Task 3} is shown in Fig.\ref{fig:evolution}. The pose error, $\mathbf{p}_e$, the force in the insertion direction, \mathbi{f}, the torque about the pitch axis, $\boldsymbol{\tau}$, and the controller parameters, $K_p^p$ and $K_p^f$, are displayed. The result shows that the initial policy failed to finish the task within 500 steps (25 s) and the learned policy completed the task at 326 steps (16.3 s). The controller parameters were tuned according to the contact force:
at the search phase after initial contact, $K_p^p$ and $K_p^f$ were kept at normal values to contact with the surface with proper contact forces; at the insertion phase, $K_p^p$ was increased to make the EEF stick to the insertion trajectory and apply force to the contact surface, and $K_p^f$ was reduced to cooperate to control the force within the upper limit.
\section{Conclusion And Future Work}
\label{section:conclusion}
In this work, we propose a framework to learn complex assembly skills from human demonstrations by using a combined trajectory and force learning approach.
The framework consists of learning the trajectory through a novel hierarchical goal-conditioned IL (HGCIL) approach and learning the force profile through an RL-based force controller.
Experimental results on simulation show that the HGCIL method can generate high-quality trajectories by imitating the demonstrated trajectories and tackling the trajectory uncertainty and the proposed framework's sample efficiency is highly enhanced compared with vanilla RL and alternative CL frameworks using BC and GCBC-relabeling as the trajectory learning methods. And experimental results on real hardware additionally show the effectiveness of the framework to quickly learn control policies of assembly tasks with varying trajectories and force requirements in the real world.






\section*{ACKNOWLEDGMENT}
The first author would like to acknowledge the financial supports from the China Scholarship Council Postgraduate Scholarship Grant 201806120019.

\bibliographystyle{IEEEtran}

\begin{thebibliography}{10}
    \providecommand{\url}[1]{#1}
    \csname url@samestyle\endcsname
    \providecommand{\newblock}{\relax}
    \providecommand{\bibinfo}[2]{#2}
    \providecommand{\BIBentrySTDinterwordspacing}{\spaceskip=0pt\relax}
    \providecommand{\BIBentryALTinterwordstretchfactor}{4}
    \providecommand{\BIBentryALTinterwordspacing}{\spaceskip=\fontdimen2\font plus
    \BIBentryALTinterwordstretchfactor\fontdimen3\font minus
      \fontdimen4\font\relax}
    \providecommand{\BIBforeignlanguage}[2]{{%
    \expandafter\ifx\csname l@#1\endcsname\relax
    \typeout{** WARNING: IEEEtran.bst: No hyphenation pattern has been}%
    \typeout{** loaded for the language `#1'. Using the pattern for}%
    \typeout{** the default language instead.}%
    \else
    \language=\csname l@#1\endcsname
    \fi
    #2}}
    \providecommand{\BIBdecl}{\relax}
    \BIBdecl
    
    \bibitem{bry2011rapidly}
    A.~Bry and N.~Roy, ``Rapidly-exploring random belief trees for motion planning
      under uncertainty,'' in \emph{2011 IEEE Int. Conf. on Robotics and
      Automation}.\hskip 1em plus 0.5em minus 0.4em\relax IEEE, 2011, pp. 723--730.
    
    \bibitem{wirnshofer2018robust}
    F.~Wirnshofer, P.~S. Schmitt, W.~Feiten, G.~v. Wichert, and W.~Burgard,
      ``Robust, compliant assembly via optimal belief space planning,'' in
      \emph{2018 IEEE Int. Conf. on Robotics and Automation}.\hskip 1em plus 0.5em
      minus 0.4em\relax IEEE, 2018, pp. 1--5.
    
    \bibitem{barto2003recent}
    A.~G. Barto and S.~Mahadevan, ``Recent advances in hierarchical reinforcement
      learning,'' \emph{Discrete event dynamic systems}, vol.~13, no. 1-2, pp.
      41--77, 2003.
    
    \bibitem{gupta2019relay}
    A.~Gupta, V.~Kumar, C.~Lynch, S.~Levine, and K.~Hausman, ``Relay policy
      learning: Solving long-horizon tasks via imitation and reinforcement
      learning,'' in \emph{Conf. on Robot Learning (CoRL)}, 2019.
    
    \bibitem{ross2010efficient}
    S.~Ross and D.~Bagnell, ``Efficient reductions for imitation learning,'' in
      \emph{Int. Conf. on artificial intelligence and statistics}, 2010, pp.
      661--668.
    
    \bibitem{ross2011reduction}
    S.~Ross, G.~Gordon, and D.~Bagnell, ``A reduction of imitation learning and
      structured prediction to no-regret online learning,'' in \emph{Int. Conf. on
      artificial intelligence and statistics}, 2011, pp. 627--635.
    
    \bibitem{cheng2018fast}
    C.-A. Cheng, X.~Yan, N.~Wagener, and B.~Boots, ``Fast policy learning through
      imitation and reinforcement,'' in \emph{Conf. on Uncertainty in Artificial
      Intelligence (UAI)}, 2018.
    
    \bibitem{hogan1984impedance}
    N.~Hogan, ``Impedance control: An approach to manipulation,'' in \emph{1984
      American control Conf.}\hskip 1em plus 0.5em minus 0.4em\relax IEEE, 1984,
      pp. 304--313.
    
    \bibitem{ijspeert2013dynamical}
    A.~J. Ijspeert, J.~Nakanishi, H.~Hoffmann, P.~Pastor, and S.~Schaal,
      ``Dynamical movement primitives: learning attractor models for motor
      behaviors,'' \emph{Neural Computation}, vol.~25, no.~2, pp. 328--373, 2013.
    
    \bibitem{kruger2014technologies}
    N.~Kr{\"u}ger, A.~Ude, H.~G. Petersen, B.~Nemec, L.-P. Ellekilde, T.~R.
      Savarimuthu, J.~A. Rytz, K.~Fischer, A.~G. Buch, D.~Kraft \emph{et~al.},
      ``Technologies for the fast set-up of automated assembly processes,''
      \emph{KI-K{\"u}nstliche Intelligenz}, vol.~28, no.~4, pp. 305--313, 2014.
    
    \bibitem{abu2015adaptation}
    F.~J. Abu-Dakka, B.~Nemec, J.~A. J{\o}rgensen, T.~R. Savarimuthu,
      N.~Kr{\"u}ger, and A.~Ude, ``Adaptation of manipulation skills in physical
      contact with the environment to reference force profiles,'' \emph{Autonomous
      Robots}, vol.~39, no.~2, pp. 199--217, 2015.
    
    \bibitem{savarimuthu2017teaching}
    T.~R. Savarimuthu, A.~G. Buch, C.~Schlette, N.~Wantia, J.~Ro{\ss}mann,
      D.~Mart{\'\i}nez, G.~Aleny{\`a}, C.~Torras, A.~Ude, B.~Nemec \emph{et~al.},
      ``Teaching a robot the semantics of assembly tasks,'' \emph{IEEE Trans. on
      Systems, Man, and Cybernetics: Systems}, vol.~48, no.~5, pp. 670--692, 2017.
    
    \bibitem{Roveda2017iterative}
    L.~{Roveda}, G.~{Pallucca}, N.~{Pedrocchi}, F.~{Braghin}, and L.~M. {Tosatti},
      ``Iterative learning procedure with reinforcement for high-accuracy force
      tracking in robotized tasks,'' \emph{IEEE Transactions on Industrial
      Informatics}, vol.~14, no.~4, pp. 1753--1763, 2018.
    
    \bibitem{inoue2017deep}
    T.~Inoue, G.~De~Magistris, A.~Munawar, T.~Yokoya, and R.~Tachibana, ``Deep
      reinforcement learning for high precision assembly tasks,'' in \emph{2017
      IEEE/RSJ Int. Conf. on Intelligent Robots and Systems}.\hskip 1em plus 0.5em
      minus 0.4em\relax IEEE, 2017, pp. 819--825.
    
    \bibitem{thomas2018learning}
    G.~Thomas, M.~Chien, A.~Tamar, J.~A. Ojea, and P.~Abbeel, ``Learning robotic
      assembly from cad,'' in \emph{2018 IEEE Int. Conf. on Robotics and
      Automation}.\hskip 1em plus 0.5em minus 0.4em\relax IEEE, 2018, pp. 1--9.
    
    \bibitem{johannink2019residual}
    T.~Johannink, S.~Bahl, A.~Nair, J.~Luo, A.~Kumar, M.~Loskyll, J.~A. Ojea,
      E.~Solowjow, and S.~Levine, ``Residual reinforcement learning for robot
      control,'' in \emph{2019 Int. Conf. on Robotics and Automation}.\hskip 1em
      plus 0.5em minus 0.4em\relax IEEE, 2019, pp. 6023--6029.
    
    \bibitem{kormushev2011imitation}
    P.~Kormushev, S.~Calinon, and D.~G. Caldwell, ``Imitation learning of
      positional and force skills demonstrated via kinesthetic teaching and haptic
      input,'' \emph{Advanced Robotics}, vol.~25, no.~5, pp. 581--603, 2011.
    
    \bibitem{takamatsu2007recognizing}
    J.~Takamatsu, K.~Ogawara, H.~Kimura, and K.~Ikeuchi, ``Recognizing assembly
      tasks through human demonstration,'' \emph{The Int. J. of Robotics Research},
      vol.~26, no.~7, pp. 641--659, 2007.
    
    \bibitem{tang2016teach}
    T.~Tang, H.-C. Lin, Y.~Zhao, Y.~Fan, W.~Chen, and M.~Tomizuka, ``Teach
      industrial robots peg-hole-insertion by human demonstration,'' in \emph{2016
      IEEE Int. Conf. on Advanced Intelligent Mechatronics (AIM)}.\hskip 1em plus
      0.5em minus 0.4em\relax IEEE, 2016, pp. 488--494.
    
    \bibitem{suomalainen2017geometric}
    M.~Suomalainen and V.~Kyrki, ``A geometric approach for learning compliant
      motions from demonstration,'' in \emph{2017 IEEE-RAS 17th Int. Conf. on
      Humanoid Robotics (Humanoids)}.\hskip 1em plus 0.5em minus 0.4em\relax IEEE,
      2017, pp. 783--790.
    
    \bibitem{atkeson1997robot}
    C.~G. Atkeson and S.~Schaal, ``Robot learning from demonstration,'' in
      \emph{Int. Conf. on Machine Learning}, vol.~97.\hskip 1em plus 0.5em minus
      0.4em\relax Citeseer, 1997, pp. 12--20.
    
    \bibitem{peters2008reinforcement}
    J.~Peters and S.~Schaal, ``Reinforcement learning of motor skills with policy
      gradients,'' \emph{Neural networks}, vol.~21, no.~4, pp. 682--697, 2008.
    
    \bibitem{kober2009policy}
    J.~Kober and J.~R. Peters, ``Policy search for motor primitives in robotics,''
      in \emph{Advances in neural information processing systems}, 2009, pp.
      849--856.
    
    \bibitem{vecerik2017leveraging}
    M.~Vecerik, T.~Hester, J.~Scholz, F.~Wang, O.~Pietquin, B.~Piot, N.~Heess,
      T.~Roth{\"o}rl, T.~Lampe, and M.~Riedmiller, ``Leveraging demonstrations for
      deep reinforcement learning on robotics problems with sparse rewards,'' in
      \emph{the Annual Conf. on Neural Information Processing Systems}, 2017.
    
    \bibitem{nair2018overcoming}
    A.~Nair, B.~McGrew, M.~Andrychowicz, W.~Zaremba, and P.~Abbeel, ``Overcoming
      exploration in reinforcement learning with demonstrations,'' in \emph{2018
      IEEE Int. Conf. on Robotics and Automation}.\hskip 1em plus 0.5em minus
      0.4em\relax IEEE, 2018, pp. 6292--6299.
    
    \bibitem{rajeswaran2017learning}
    A.~Rajeswaran, V.~Kumar, A.~Gupta, G.~Vezzani, J.~Schulman, E.~Todorov, and
      S.~Levine, ``Learning complex dexterous manipulation with deep reinforcement
      learning and demonstrations,'' in \emph{Robotics: Science and Systems (RSS)},
      2018.
    
    \bibitem{bain1995framework}
    M.~Bain and C.~Sammut, ``A framework for behavioural cloning.'' in
      \emph{Machine Intelligence 15}, 1995, pp. 103--129.
    
    \bibitem{kaelbling1993learning}
    L.~P. Kaelbling, ``Learning to achieve goals,'' in \emph{IJCAI}.\hskip 1em plus
      0.5em minus 0.4em\relax Citeseer, 1993, pp. 1094--1099.
    
    \bibitem{schaul2015universal}
    T.~Schaul, D.~Horgan, K.~Gregor, and D.~Silver, ``Universal value function
      approximators,'' in \emph{Int. Conf. on Machine Learning}, 2015, pp.
      1312--1320.
    
    \bibitem{ding2019goal}
    Y.~Ding, C.~Florensa, P.~Abbeel, and M.~Phielipp, ``Goal-conditioned imitation
      learning,'' in \emph{Advances in Neural Information Processing Systems},
      2019, pp. 15\,298--15\,309.
    
    \bibitem{lynch2019learning}
    C.~Lynch, M.~Khansari, T.~Xiao, V.~Kumar, J.~Tompson, S.~Levine, and
      P.~Sermanet, ``Learning latent plans from play,'' in \emph{Conf. on Robot
      Learning (CoRL)}, 2019.
    
    \bibitem{haarnoja2018soft}
    T.~Haarnoja, A.~Zhou, P.~Abbeel, and S.~Levine, ``Soft actor-critic: Off-policy
      maximum entropy deep reinforcement learning with a stochastic actor,'' in
      \emph{Int. Conf. on Machine Learning}, 2018, pp. 1861--1870.
    
    \bibitem{ota2020tf2rl}
    K.~Ota, ``Tf2rl,'' \url{https://github.com/keiohta/tf2rl/}, 2020.
    
    \bibitem{levine2016end}
    S.~Levine, C.~Finn, T.~Darrell, and P.~Abbeel, ``End-to-end training of deep
      visuomotor policies,'' \emph{The J. of Machine Learning Research}, vol.~17,
      no.~1, pp. 1334--1373, 2016.
    
    \bibitem{beltran2020learning}
    C.~C. Beltran-Hernandez, D.~Petit, I.~G. Ramirez-Alpizar, T.~Nishi, S.~Kikuchi,
      T.~Matsubara, and K.~Harada, ``Learning force control for contact-rich
      manipulation tasks with rigid position-controlled robots,'' \emph{IEEE
      Robotics and Automation Letters}, vol.~5, no.~4, pp. 5709--5716, 2020.
    
    \bibitem{bellegarda2019training}
    G.~{Bellegarda} and K.~{Byl}, ``Training in task space to speed up and guide
      reinforcement learning,'' in \emph{2019 IEEE/RSJ Int. Conf. on Intelligent
      Robots and Systems}, 2019, pp. 2693--2699.
    
    \bibitem{martin2019variable}
    R.~{Martín-Martín}, M.~A. {Lee}, R.~{Gardner}, S.~{Savarese}, J.~{Bohg}, and
      A.~{Garg}, ``Variable impedance control in end-effector space: An action
      space for reinforcement learning in contact-rich tasks,'' in \emph{2019
      IEEE/RSJ Int. Conf. on Intelligent Robots and Systems}, 2019, pp. 1010--1017.

\end{thebibliography}

\end{document}